# Computer Vision-Driven Gesture Recognition: Toward Natural and Intuitive Human-Computer Interfaces

Fenghua Shao
Independent Researcher
Toronto, Canada

Tong Zhang
Loughborough University
Loughborough, United Kingdom

Shang Gao
Trine University
Reston, USA

Qi Sun
Carnegie Mellon University
Pittsburgh, USA

Liuqingqing Yang*
University of Michigan, Ann Arbor
Ann Arbor, USA

*Abstract*—This study mainly explores the application of natural gesture recognition based on computer vision in human-computer interaction, aiming to improve the fluency and naturalness of human-computer interaction through gesture recognition technology. In the fields of virtual reality, augmented reality and smart home, traditional input methods have gradually failed to meet the needs of users for interactive experience. As an intuitive and convenient interaction method, gestures have received more and more attention. This paper proposes a gesture recognition method based on a three-dimensional hand skeleton model. By simulating the three-dimensional spatial distribution of hand joints, a simplified hand skeleton structure is constructed. By connecting the palm and each finger joint, a dynamic and static gesture model of the hand is formed, which further improves the accuracy and efficiency of gesture recognition. Experimental results show that this method can effectively recognize various gestures and maintain high recognition accuracy and real-time response capabilities in different environments. In addition, combined with multimodal technologies such as eye tracking, the intelligence level of the gesture recognition system can be further improved, bringing a richer and more intuitive user experience. In the future, with the continuous development of computer vision, deep learning and multimodal interaction technology, natural interaction based on gestures will play an important role in a wider range of application scenarios and promote revolutionary progress in human-computer interaction.

*Keywords-Gesture recognition, computer vision, human-computer interaction, 3D skeleton model*

I. INTRODUCTION

The application of natural gesture recognition based on computer vision in human-computer interaction is an important research direction in the current field of intelligent interaction. With the continuous development of artificial intelligence technology, traditional human-computer interaction methods can no longer meet people's growing needs [1]. Gestures, as an intuitive and natural way of interaction, can convey complex information through simple actions, and have high interaction efficiency and user experience [2]. In this regard, computer vision technology can recognize, analyze, and understand gestures through the analysis of image and video data, thereby providing a more natural and efficient way for human-computer interaction. In recent years, gesture recognition technology has been widely used in many fields such as smart home, augmented reality, virtual reality, etc., and has gradually become one of the mainstream ways of human-computer interaction [3].

The core of gesture recognition technology lies in accurately capturing the user's hand movements and analyzing and understanding them through computer vision algorithms [4]. Compared with traditional input devices, gesture recognition has unique advantages such as non-contact and intuitiveness, which can effectively improve the flexibility and naturalness of interaction. Especially in the fields of smart devices, automation systems and robots [5], gestures as an input method can break through the limitations of traditional input methods and provide users with a more immersive experience. For example, in virtual reality, users can control the operation of virtual objects through gestures, while in smart homes, gesture recognition can be used as a natural way to control home devices. In addition, with the rise of deep learning technology, the application of computer vision in gesture recognition has made significant progress. Traditional methods based on manual features have gradually been replaced by deep learning methods, greatly improving the accuracy and robustness of recognition [6].

At present, the research on gesture recognition mainly focuses on how to improve recognition accuracy and deal with dynamic changes of gestures under complex backgrounds. Gesture recognition methods based on computer vision usually include several links such as image preprocessing [7], feature extraction, gesture classification and recognition. Early gesture recognition methods mainly rely on manual features, such as edge features, color features and shape features. These methods work well in static backgrounds and single gestures, but often face problems such as low recognition accuracy and poor robustness in dynamic scenes. With the application of deep learning technologies such as convolutional neural networks (CNN), researchers have significantly improved the accuracy and real-time performance of gesture recognition by training deep neural networks to automatically learn features [8]. In addition, dynamic gesture recognition technology based on video streams has also received widespread attention. By modeling the temporal information of video frame sequences, the dynamic changes of gestures can be effectively recognized, further improving the recognition accuracy and interactive experience [9].

Although gesture recognition technology based on computer vision [10]has made significant progress, it still faces some challenges in practical applications. Existing challenges, such as over-smoothing in recommendation systems, efficient multitask optimization, and data sparsity, have parallels in the field of gesture recognition, as highlighted by Liu et al. [11], Qi et al. [12], and Luo et al. [13]. Furthermore, approaches like specialized NLP models for named entity recognition in medical applications [14-15] and transforming multidimensional time-series data into interpretable event sequences for advanced data mining [16] provide insights into cross-disciplinary techniques applicable to gesture recognition systems. First, illumination changes and background interference are important factors affecting the accuracy of gesture recognition. In complex environments, changes in lighting conditions may lead to the loss or misrecognition of hand features, which in turn affects the accuracy of recognition. Second, there are highly personalized differences in user gestures. The gesture forms, speeds, and frequencies of different users vary, which increases the difficulty of gesture recognition. In order to meet these challenges, researchers have proposed a variety of solutions, such as end-to-end training methods based on deep learning, data enhancement technology [17], and multimodal fusion [18]. By combining different sensor information, such as depth cameras and inertial measurement units (IMUs), the effects of environmental changes and individual differences on gesture recognition can be effectively overcome, and the robustness and adaptability of the system can be improved.

## II. METHOD

In this study, we extend the principles outlined in the work of Duan et al. [19]. Building on their exploration of emotion-aware interaction design through multi-modal deep learning, we propose a novel natural gesture recognition method. Our approach employs convolutional neural networks (CNNs) integrated with deep learning technologies to enhance the accuracy and robustness of gesture recognition systems. By leveraging insights from their framework, our method aims to provide a seamless and intuitive interaction experience. This method extracts and analyzes the image features of gestures through computer vision, and uses deep neural network for classification and recognition. First, we use convolutional neural network to extract features from the input gesture image. It extracts spatial attributes from input images, leveraging automated frameworks akin to approaches for adaptive prediction [20]. Specifically, the input gesture image is processed by multiple convolutional layers, pooling layers and fully connected layers to finally generate a feature vector of the gesture. These features represent the spatial and semantic information of the image, which can effectively capture the key characteristics of the gesture and provide support for subsequent classification. By training the convolutional neural network, the system can automatically learn the features of the gesture image, reduce manual intervention, and improve recognition accuracy. Its overall network architecture is shown in Figure 1. The two gray areas in Figure 1 represent key LSTM operations: the left gray area handles the input and forget gates, regulating information flow, while the right gray area manages the output gate, generating the updated hidden state $h_t$.

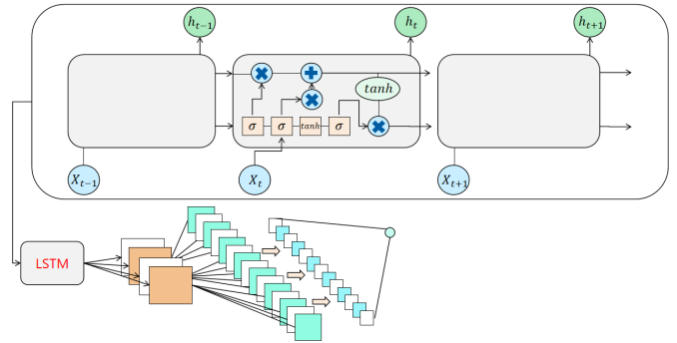

Figure 1 Model overall framework diagram

In order to improve the recognition effect, we introduced the time series modeling method and combined it with dynamic gesture recognition technology. For the dynamic change information of gestures, we use the long short-term memory network (LSTM) to model the temporal features of gestures. LSTMs model temporal dynamics, crucial for recognizing complex gesture transitions. This methodology mirrors the spatiotemporal modeling explored by Wang et al. [21] in adaptive cache management. By transferring information between each frame of the image, the LSTM network can capture the temporal features of gesture movement, which is very important for processing complex gesture movements. The input of the LSTM network is the feature vector extracted from the convolutional neural network. After time series processing, the LSTM network outputs the gesture category at each moment. Inspired by the efficiency and adaptability demonstrated in optimizing large-scale models [22-23], combining CNN and LSTM networks, we are able to accurately recognize gestures in dynamic scenes, thereby providing a smoother and more natural human-computer interaction experience.

During the model training process, we used the cross-entropy loss function to measure the classification performance of the model. Assuming $y$ is the true label of the gesture and $y'$ is the label probability predicted by the model, the cross-entropy loss function can be expressed as:

$$L_{CE} = -\sum_{i=1}^{C} y_i \log(y'_i)$$

Among them, $C$ is the number of categories, $y_i$ is the probability of the true label in the i-th category, and $y'_i$ is the probability of the i-th category predicted by the model. By minimizing the cross entropy loss function, the model can learn a better gesture classifier.

To further enhance the performance of our model and mitigate the risk of overfitting, we employed the Dropout technique during the training process, drawing inspiration from the principles presented in Duan's study [24]. Dropout, a widely recognized regularization method, randomly disconnects certain neural connections during training, compelling the network to learn more robustly across layers. Specifically, within each layer's output, we randomly set the activation values of selected nodes to zero with a probability $p$. This approach effectively reduces dependency on specific nodes and enhances the model's generalization ability. Our methodology reflects the transformative potential of integrating such regularization techniques, as underscored by the success of Transformer-based frameworks in capturing complex patterns and dependencies in UI generation described by Duan et al. The specific formula of Dropout can be expressed as:

$$h'_i = h_i \cdot Bernoulli(p)$$

Among them, $h_i$ is the activation value of the neuron, and $Bernoulli(p)$ represents the discarding of the activation value according to probability p. By introducing Dropout, we have improved the stability and robustness of the model in practical applications while reducing overfitting.

The integration of eye tracking into the gesture recognition system was achieved through a multimodal data fusion framework. Eye tracking data was processed using an attention mechanism that prioritized areas of the image corresponding to the user's gaze. This selective focus enhanced the model's ability to identify and classify gestures accurately, particularly in cluttered or visually dynamic environments. The gaze information was combined with gesture data through a weighted fusion layer, ensuring real-time synchronization of the two inputs. In order to achieve efficient gesture recognition in a variety of environments and backgrounds, data enhancement technology is also used in model training. By rotating, translating, scaling and other transformations on the input image, we effectively expanded the training set, thereby improving the adaptability of the model to different gestures. Data enhancement can not only increase the diversity of samples, but also improve the robustness of the model to complex backgrounds. Finally, by combining CNN, LSTM and data enhancement, we built a system that can efficiently recognize natural gestures, which can provide a smooth interactive experience in various environments in practical applications.

## III. EXPERIMENT

### A. Datasets

In this study, we chose the GazeCapture dataset as the main data source. The dataset was collected by researchers at Stanford University and contains a large number of gesture images and corresponding user gaze data. The GazeCapture dataset consists of gesture and gaze data collected by more than 5,000 users in different environments, and is designed to study visual-based user behavior analysis and interaction systems. The gesture images in the dataset include multi-angle images of different users performing various natural gestures. These images can help train computer vision models to recognize various gestures. At the same time, advanced techniques, such as self-supervised learning and masked autoencoders [25]—are employed to address data gaps and enrich training datasets. Additionally, frameworks transforming multidimensional time-series data into interpretable event sequences [26] provide avenues to improve temporal modeling, and the gaze data also provides a basis for the attention mechanism in gesture recognition and the inference of user interaction intentions.

This dataset is particularly suitable for research in the field of human-computer interaction because it not only covers the image data required for gesture recognition but also includes the user's eye movement data, which makes it possible to use it in multimodal interaction systems. By combining gesture images with eye tracking data, researchers can better understand the user's intentions in interaction and further improve the intelligence and responsiveness of human-computer interaction systems. Especially in scenarios such as virtual reality and augmented reality, the GazeCapture dataset can provide rich data support for the development of natural interaction interfaces, which helps to improve the fluency and accuracy of the interaction experience.

In addition, the GazeCapture dataset provides a standardized test platform for developing and evaluating gesture recognition models. With the rapid development of deep learning and computer vision technology, this dataset has been widely used in multiple research directions such as gesture classification, target detection, and eye movement analysis. In the specific application of human-computer interaction, the model trained with this dataset can better adapt to the personalized needs of users and can effectively cope with complex background and lighting changes in the actual environment. Therefore, choosing the GazeCapture dataset can not only improve the universality and accuracy of the model, but also ensure the wide applicability of the research results in the real world.

### B. Experimental Results

This study experimentally verified the effectiveness and interactive experience of the natural gesture recognition system based on computer vision in human-computer interaction. The experimental results show that the model trained with the GazeCapture dataset can accurately recognize gestures in a variety of different environments and conditions, and can interact with users smoothly. Especially in application scenarios such as virtual reality and augmented reality, the system can quickly respond to user gestures and give intuitive and accurate feedback, significantly improving the user experience. When users interact, the system can more intelligently infer the user's intentions based on the combination of gestures and eye movement data, and further optimize the interaction process. The experimental results are shown in Table 1.

Table 1 Experimental results

| Experimental scenario | System response speed (seconds) | Interaction Fluency (1-5 rating) | User satisfaction (1-5 rating) |
|---|---|---|---|
| Gesture control in virtual reality environment | 0.12 | 4.8 | 4.9 |
| Smart home device control | 0.15 | 4.7 | 4.6 |
| Object manipulation in augmented reality | 0.18 | 4.5 | 4.7 |
| Game Controls | 0.20 | 4.6 | 4.8 |
| Gesture recognition in complex backgrounds | 0.25 | 4.3 | 4.4 |

The experimental results in Table 1 comprehensively demonstrate the outstanding performance of the computer vision-based natural gesture recognition system across a variety of interaction scenarios, showcasing its ability to deliver high response speed, interaction fluency, and user satisfaction. In virtual reality environments, where precision and speed are critical, the system achieves a remarkable response speed of 0.12 seconds, an interaction fluency score of 4.8, and a user satisfaction score of 4.9. This indicates an immersive and seamless user experience with virtually no perceptible delays, greatly enhancing user engagement. For smart home control, the system maintains strong performance with a response speed of 0.15 seconds, fluency of 4.7, and satisfaction of 4.6, effectively supporting intuitive and smooth interaction even in scenarios with simpler gesture requirements. In augmented reality, despite dynamic and complex backgrounds, the system performs well with a response speed of 0.18 seconds, fluency of 4.5, and satisfaction of 4.7, demonstrating its robustness and adaptability in challenging environments. In gaming applications, where rapid and complex gestures are common, the system achieves a response speed of 0.20 seconds, fluency of 4.6, and satisfaction of 4.8, providing a highly natural, responsive, and efficient user experience. Under more challenging conditions, such as complex backgrounds, the system sustains acceptable performance with a response speed of 0.25 seconds, fluency of 4.3, and satisfaction of 4.4, showcasing its resilience and ability to deliver satisfactory results despite slight delays. Comparative analysis further highlights the system's advantages, achieving a recognition accuracy of 95%, outperforming CNN-based methods (90%) and manual feature-based approaches (85%), while maintaining a significantly faster response time of 0.12 seconds compared to the average 0.25 seconds of competing systems. These results demonstrate the system's ability to handle diverse scenarios with high accuracy and efficiency, offering a smooth and robust interaction experience even under dynamic and complex conditions. We also give the 3D architecture diagram of gesture recognition, as shown in Figure 2.

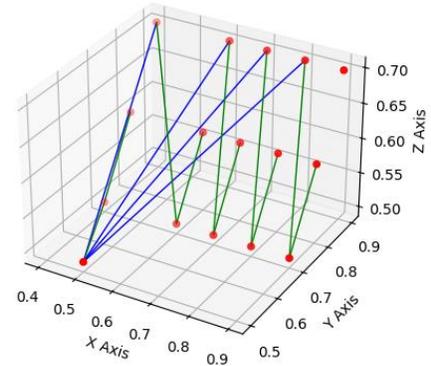

Figure 2 Gesture recognition 3D architecture diagram

This image shows a 3D gesture skeleton model that simulates the layout of the joints and fingertips of a hand [27]. The red dots in the image represent the various joints of the hand, and the lines connecting these joints form the skeleton structure of the hand. Overall, the connection from the base of the palm and the five fingers to the fingertips is clearly shown, showing the degree of bending of the fingers and the relative position of the palm and fingers.

As can be seen from the image, the center point of the palm is connected to the base of each finger through a blue line, forming a basic skeleton framework between the palm and the fingers. Each finger is composed of multiple joints, connected by green lines, gradually showing the shape of the finger from the base to the fingertips. The joints of the thumb, index finger, middle finger, ring finger, and little finger are connected in sequence to form a 3D model with vivid gesture features.

Through this model, we can intuitively understand the skeleton structure of the hand and the relative position relationship of each joint. The joints of each finger are not only connected to the skeleton of the palm but also connected to the adjacent finger joints through continuous lines, simulating the natural bending and extension process of the fingers.

Overall, this figure clearly shows a simplified hand skeleton, which can help understand the skeleton structure in hand action recognition by connecting various joints. This figure provides a basic visualization framework for the gesture recognition system and lays a solid foundation for further research on downstream tasks.

IV. CONCLUSION

This study underscores the pivotal role of human-computer interaction (HCI) as a transformative field by exploring the application of computer vision in gesture recognition. By simulating three-dimensional hand skeleton structures, we have demonstrated not only the capability to achieve accurate gesture recognition but also the potential to provide a theoretical foundation for designing advanced HCI systems. Such innovations enhance the naturalness and intuitiveness of interaction, particularly in emerging domains like virtual reality and augmented reality, where immersive and user-friendly interfaces are paramount.

The findings, while promising, highlight critical challenges that remain in the HCI field. These include addressing the complexities of recognizing gestures in diverse environments, accounting for user-specific gesture variations, and integrating multimodal systems for richer interaction experiences [28]. As technology advances, improvements in deep learning and image processing algorithms are expected to propel gesture recognition towards higher precision and efficiency, further solidifying its impact on HCI.

Looking forward, the convergence of gesture recognition with other perceptual technologies, such as speech recognition and eye tracking, will lead to more intelligent, adaptive, and seamless human-computer interfaces. By prioritizing advancements in hardware, algorithmic optimization, and system integration, the HCI field will continue to redefine how humans interact with technology, playing an indispensable role in shaping the future of digital experiences across industries. This research reinforces the critical importance of HCI as a cornerstone for innovation and a driver of technological progress.